\documentclass[draft]{agujournal2019}
\usepackage{url} 
\usepackage{lineno}
\usepackage[percent]{overpic}

\usepackage{xcolor}

\usepackage{amsmath} 
\usepackage{amssymb} 
\usepackage{booktabs}
\usepackage{natbib}
\usepackage{graphicx}
\usepackage{booktabs} 
\usepackage{siunitx} 
\usepackage{adjustbox} 
\usepackage{subcaption} 

\draftfalse

\journalname{JGR: Machine Learning and Computation}
\begin{document}

\title{Observation-driven correction of numerical weather prediction for marine winds}

\authors{Matteo Peduto\affil{1,2}, Qidong Yang\affil{1}, Jonathan Giezendanner\affil{1}, Devis Tuia\affil{2}, and Sherrie Wang\affil{1}}

\affiliation{1}{Massachusetts Institute of Technology, Cambridge, MA, USA}
\affiliation{2}{\'Ecole Polytechnique F\'ed\'erale de Lausanne (EPFL), Lausanne, Switzerland}

 
\begin{keypoints}
\item ORCA, a transformer, corrects NWP marine winds by assimilating irregular, time-varying in-situ observations via set-based attention.
\item ORCA reduces forecast errors by 45\% at one-hour lead time and 13\% at 48 hours across all observation platform types.
\item Single pass inference at arbitrary coordinates enables both site-specific and basin scale predictions suitable for operational workflows.
\end{keypoints}

\begin{abstract}
Accurate marine wind forecasts are essential for safe navigation, ship routing, and energy operations, yet they remain challenging because observations over the ocean are sparse, heterogeneous, and temporally variable. We present an observation-informed correction approach for global numerical weather prediction (NWP) of marine winds. Rather than forecasting winds directly, we learn local correction patterns by assimilating the latest in-situ observations to adjust the Global Forecast System (GFS) output. We propose ORCA (Observation-informed Real-time Correction with Attention), a transformer-based deep learning architecture that (i) handles irregular and time-varying observation sets through masking and set-based attention mechanisms, (ii) conditions predictions on recent observation--forecast pairs via cross-attention, and (iii) employs cyclical time embeddings and coordinate-aware location representations to enable single-pass inference at arbitrary spatial coordinates.
We evaluate ORCA over the Atlantic Ocean using observations from the International Comprehensive Ocean-Atmosphere Data Set (ICOADS) as reference. ORCA reduces GFS 10-meter wind error at all lead times up to 48 hours, achieving 45\% improvement at 1-hour lead time and 13\% improvement at 48-hour lead time. Spatial analyses reveal the most persistent improvements along coastlines and shipping routes, where observations are most abundant. The tokenized architecture naturally accommodates heterogeneous observing platforms (ships, buoys, tide gauges, and coastal stations) and produces both site-specific predictions and basin-scale gridded products in a single forward pass.
These results demonstrate a practical, low-latency post-processing approach that complements NWP by learning to correct systematic forecast errors.
\end{abstract}

\section*{Plain Language Summary}
Accurate wind forecasts over the open ocean are critical for maritime navigation, ship routing, offshore energy operations, and storm warnings. However, the ocean has far fewer weather observation stations than land areas, causing global weather prediction models to reproduce local conditions inaccurately. We adopt a practical post-processing approach: rather than replacing an established global forecast model (the Global Forecast System), we correct its predictions using recent measurements from ships, buoys, tide gauges, and coastal weather stations. We trained a machine-learning model based on transformer architecture to learn when and how to adjust forecasted winds based on these observations, their geographic locations, and temporal information including time of day and season. When tested over the Atlantic Ocean, our method reduces forecast errors at all prediction lead times up to 48 hours. The model generates predictions rapidly at any location, making it well-suited for operational forecasting workflows. This work demonstrates how integrating global weather models with observational data and artificial intelligence can produce more reliable marine wind forecasts.

\section{Introduction}
 
\begin{figure}[htbp]
 \centering

 \begin{subfigure}[t]{\linewidth}
 \centering
 \adjustbox{width=\linewidth,totalheight=6cm,keepaspectratio}{%
 \begin{overpic}[width=\linewidth]{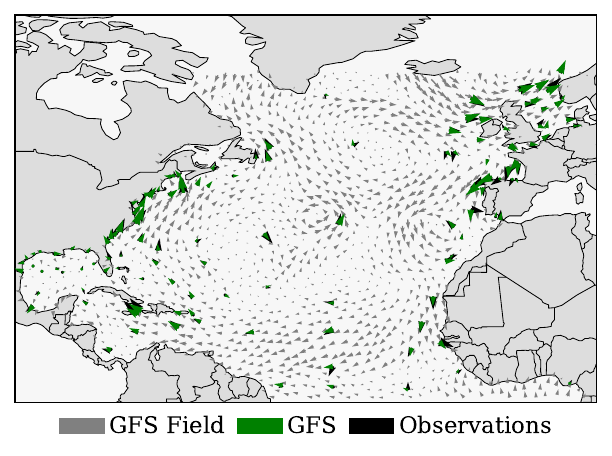}
 \put(4,52){%
 \includegraphics[width=0.18\linewidth]{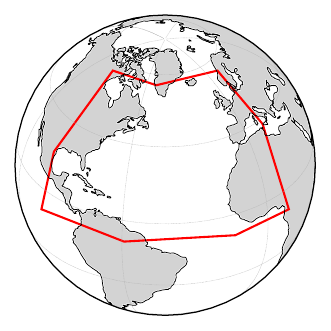}
 }
 \end{overpic}
 }
 \caption{Wind vectors over the North Atlantic. Grey arrows indicate the GFS forecast field, black arrows show in-situ observations, and green arrows represent GFS predictions corresponding to observation locations. The inset map outlines the analysis domain. Discrepancies between green and black arrows indicate locations where GFS deviates from observations, illustrating the systematic biases that the correction model targets.}
 \label{fig:panel_a}
 \end{subfigure}

 \vspace{0.5cm} 

 \begin{subfigure}[t]{\linewidth}
 \centering
 \adjustbox{width=\linewidth,totalheight=6cm,keepaspectratio}{%
 \includegraphics[width=\linewidth]{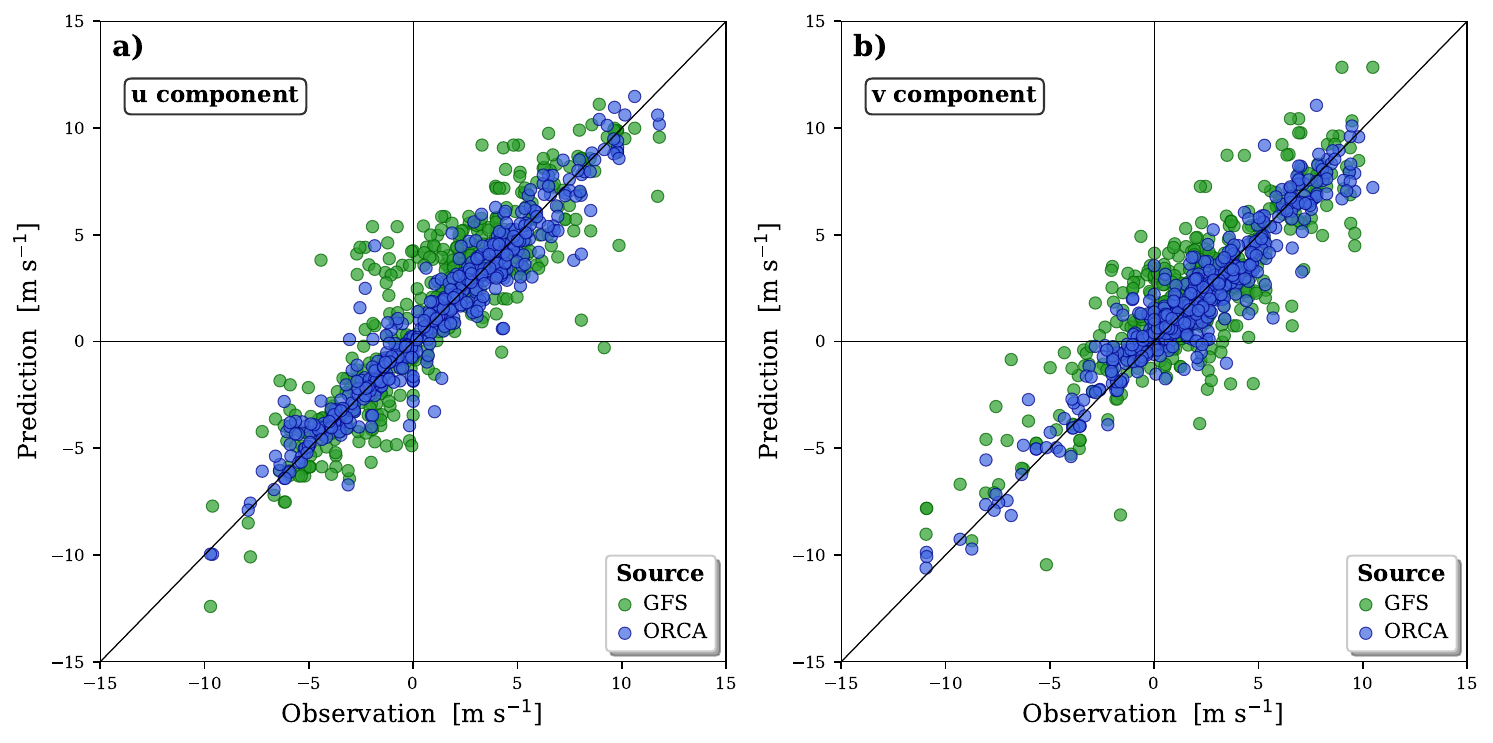}
 }
 \caption{Comparison between observed and predicted wind components at ICOADS observation locations shown in panel~(a). Green symbols denote GFS forecasts and blue symbols ORCA outputs. The left sub-panel shows the $u$~component and the right sub-panel the $v$~component; the 1:1 line marks perfect agreement.}
 \label{fig:panel_c}
 \end{subfigure}

 \caption{Comparison of spatial and statistical wind patterns from GFS forecasts, in-situ observations, and ORCA predictions. Panel~(a) shows the spatial wind field; panel~(b) compares predicted and observed wind components at observation sites.}
 \label{fig:quiver}
\end{figure}

Accurate marine wind forecasts are essential for navigation safety, ship routing, offshore energy operations, and hazard early warning, yet they remain challenging because observations over the ocean are sparse, heterogeneous, and unevenly distributed in space and time. Global numerical weather prediction (NWP) systems provide coherent large-scale dynamics, but contain biases at many locations (Figure \ref{fig:quiver}).
 
Prior work on NWP bias correction -- from classical statistical methods to recent deep learning approaches -- has demonstrated substantial improvements for surface weather variables, including wind~\citep{glahn1972mos,rasp2018nnpost,NOAA_ML4BC_2025}. However, these methods generally assume fixed observation locations or regular grids. Over the ocean, the set of available observations changes from hour to hour as ships move, buoys go offline, and reporting intervals vary across platforms. No current framework jointly handles this irregular, time-varying observation geometry while enabling inference at arbitrary coordinates -- two requirements that are essential for operational marine wind correction (Section~\ref{sec:related_work}).

In this work, we propose a correction-based approach to marine wind forecasting: rather than predicting winds directly, we learn local patterns using the in-situ observations to adjust NWP such as NOAA's Global Forecast System (GFS). The central idea is to preserve the large-scale flow provided by NWP while applying localized, observation-informed corrections wherever and whenever observations are available. 
We design a transformer-based architecture that assimilates irregular, time-varying sets of marine observations from heterogeneous platforms (ships, buoys, tide gauges, coastal stations) and produces single-pass predictions at arbitrary locations, enabling both gridded and site-specific products suitable for operational workflows.
 
We evaluate our approach over the Atlantic Ocean using observations from the International Comprehensive Ocean-Atmosphere Data Set (ICOADS) as a reference (Section~\ref{sec:icoads}), with GFS, the European Center for Medium-Range Weather Forecasts Reanalysis (ERA5), and a MOS-style linear regression \citep{glahn1972mos} as baseline models. Spatial analyses show most consistent improvements along coasts and shipping routes, where observations are denser.

 
Our contributions are as follows.
\begin{enumerate}
 
 \item 
 
 \textbf{Architecture for irregular marine observation geometry.} We propose a transformer-based model that treats each forecast time step as a variable-size set of heterogeneous marine reports. Self-attention summarizes recent observation-forecast pairs regardless of their number or spatial distribution, and cross-attention conditions corrections at any target location on this dynamic context. Unlike prior NWP post-processing architectures that assume fixed grids or station networks, ours adapts to whichever platforms report at inference time without retraining or interpolation (Section~\ref{sec:methods_model}).
 
 \item 
 
 \textbf{Scalable arbitrary-location inference.} The architecture produces single-pass predictions at any coordinates---from individual platforms to basin-scale grids---without fixed neighborhoods or interpolation, supporting both site-specific and gridded products within the same model (Sections~\ref{sec:experimental} and \ref{sec:results}).
 
 \item 
 
 \textbf{Empirical validation across the Atlantic.} Applied to 10-m wind forecasting over the entire Atlantic Ocean, the model achieves a \textbf{45\% reduction in GFS mean vector wind error at 1-hour lead time} and \textbf{13\% at 48 hours}, systematically improving predictions across all observation platforms---including ships, buoys, tide gauges, and coastal stations (Sections~\ref{sec:experimental} and \ref{sec:results}).
 
\end{enumerate}

\section{Related Work}
\label{sec:related_work}
\noindent\textbf{Modern weather forecasting couples dynamical models with statistical post-processing to correct systematic errors.}
Since the 1970s, Model Output Statistics (MOS) and its probabilistic successors have provided principled ways to calibrate NWP outputs against observations \citep{glahn1972mos,gneiting2005emos,raftery2005bma,wilks2011statistical}. These approaches address mean and distributional biases but are limited in representing nonlinear, state-dependent, and spatially varying errors. Deep post-processing methods have therefore gained traction: neural networks trained on ensemble and auxiliary predictors improve calibration and sharpness over classical regression \citep{rasp2018nnpost,scheuerer2020qrforests}, and recent applications show ML-based bias correction enhancing GFS precipitation and near-surface temperature forecasts in real-world settings \citep{sun2023gfs_precip,NOAA_ML4BC_2025}. However, these deep post-processing approaches generally assume a fixed set of predictor locations and do not account for observation sets that change in composition and spatial coverage from one forecast cycle to the next---a defining characteristic of the marine environment.

\noindent\textbf{End-to-end deep learning has rapidly advanced from nowcasting to global medium-range forecasting with near-instant inference.}
Data-driven global models such as FourCastNet, GraphCast, and Pangu-Weather demonstrate medium-range skill competitive with leading operational systems while achieving orders-of-magnitude speedups at $\sim$0.25$^{\circ}$ resolution \citep{pathak2022fourcastnet,lam2023graphcast,bi2023pangu}. For short-range, high-resolution targets, the MetNet family delivers kilometer-scale, minute-level updates for multiple surface variables, including wind \citep{metnet3_arxiv,metnet3_blog}. These developments make large ensembles and on-demand forecasts at arbitrary coordinates computationally feasible. However, these global models are trained and evaluated on regular grids and do not ingest real-time in-situ observations to correct local biases at specific offshore locations. While these models motivate key architectural choices such as efficient attention and scalable inference, our application targets short-range (1--48\,h) regional correction over the Atlantic rather than global medium-range forecasting.

\noindent\textbf{Marine wind forecasting is uniquely constrained by sparse, heterogeneous observations and marine-specific model errors.}
Compared to land, the ocean is observationally sparse and relies on a mix of buoys, ships, coastal stations, and satellite surface winds \citep{atlas2011ccmp,ccmp_remss}.
Data gaps degrade initialization and can leave persistent marine wind biases in dynamical models; targeted assimilation has shown measurable improvements, including for near-surface winds over the tropics and high latitudes \citep{rennie2021aeolus,ecmwf_aeolus_impact,zuo2023aeolus}. Physics-guided and hybrid ML systems that fuse NWP with local sensors improve site-specific offshore wind forecasts across heights and lead times \citep{airu_wrf,deepmide_2024}. Concurrently, deep spatiotemporal models (LSTM/GNN/attention) trained on buoy networks and coastal sensors reduce error relative to raw NWP and classical baselines \citep{han2023offshore_dnn,liu2023agn_gnn,dong2024marine_gnn}. A common limitation across these studies is their reliance on fixed station sets: models are trained and evaluated on specific buoy or coastal networks and cannot generalize to locations or platform configurations not seen during training.

\noindent\textbf{Learning from irregular, time-varying sets of observations benefits from attention architectures and geospatial encodings.}
Set-aware transformers and Perceiver-style cross-attention provide permutation-invariant processing of variable-length inputs and flexible conditioning from one set (recent observations) onto another (targets) \citep{lee2019set,jaegle2021perceiverio}. To query forecasts at arbitrary points without gridding artifacts, geospatial encoders that respect spherical geometry yield globally consistent coordinate features for transformers \citep{russwurm2023locationencoder,tancik2020fourier,sitzmann2020siren}. These components have been validated independently, but have not yet been combined into a single post-processing framework that conditions on a dynamic set of marine observations and predicts at arbitrary ocean coordinates.


\noindent\textbf{Summary and gap.}
In summary, no existing approach jointly addresses the three requirements of operational marine wind correction: (i) assimilating irregular, time-varying multi-platform observations, (ii)~learning state-dependent NWP error patterns conditioned on the most recent available data, and (iii)~producing corrections at arbitrary ocean coordinates in a single forward pass. Machine learning weather models routinely produce global fields in under a minute and support dense spatiotemporal sampling \citep{lam2023graphcast,bi2023pangu,pathak2022fourcastnet}, and set-based attention and spherical location encoders have each proven effective in isolation. Our model combines these elements---set-based self-attention over dynamic observation--forecast pairs, cross-attention conditioning from context to targets, and learned spherical location encoders---into a unified post-processing framework for 10-m winds over the Atlantic, demonstrating a practical approach that scales to arbitrary target sets while preserving and improving global NWP predictions.

\section{Data}
\label{sec:data}

Our study combines global reanalysis, operational forecasts, and in situ marine observations to evaluate and train the proposed models. 
Our analysis focuses on the North Atlantic domain ($2^\circ$--$64^\circ$N, $98^\circ$W--$11.5^\circ$E) over the period April~2015 to September~2024. 
We used three complementary datasets: the International Comprehensive Ocean Atmosphere Data Set (ICOADS) as the observational reference, the ECMWF Reanalysis v5 (ERA5) as a physically consistent benchmark, and the Global Forecast System (GFS) as the operational forecast baseline.

\subsection{International Comprehensive Ocean--Atmosphere Data Set (ICOADS)}
\label{sec:icoads}

\begin{figure}[h]
 \centering
 \includegraphics[width=\linewidth]{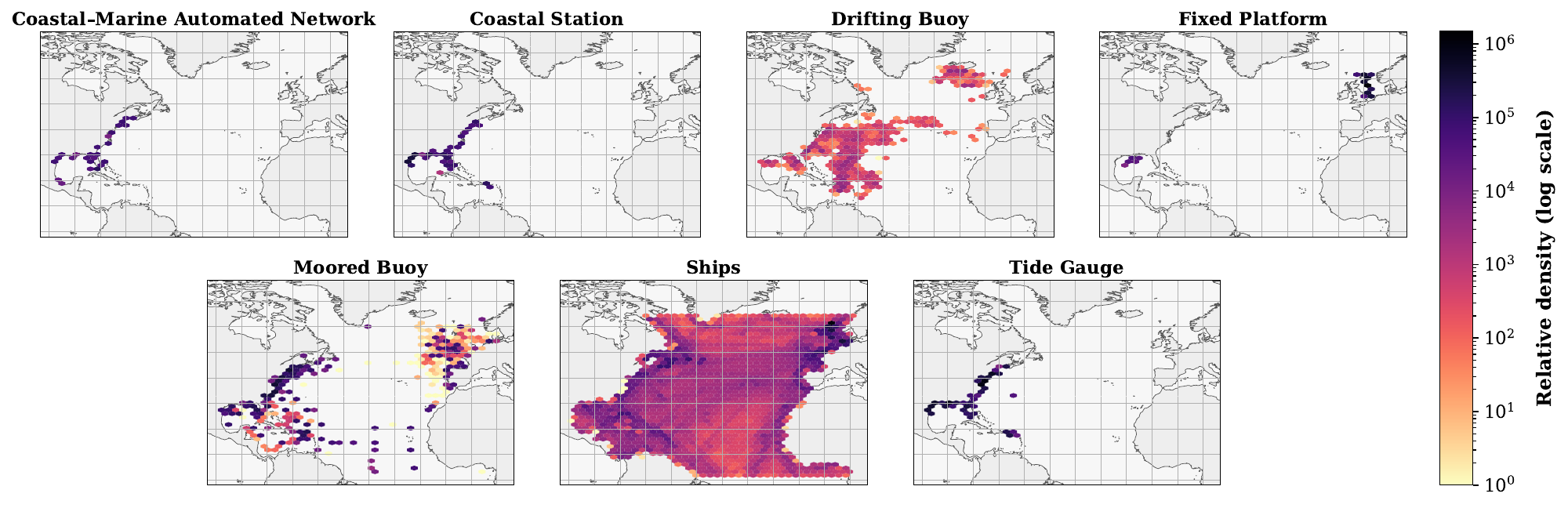}
 \caption{Spatial density of ICOADS observations by platform type (2015--2024). Higher densities occur near coastlines, while ship observations dominate the open ocean.}
 \label{fig:data_split}
\end{figure}

 ICOADS provides the most extensive collection of global marine meteorological and ocean surface observations. It aggregates reports from ships, buoys, coastal and Coastal-Marine Automated Network (C-MAN) stations, and other near-surface platforms, harmonized through quality flags and hourly timestamp rounding. After filtering, the dataset includes 6,625 unique platforms during the study period, with an average of approximately 70 measurements per platform. Ships, moored buoys, and tide gauges record the majority of observations in ICOADS, while drifting buoys contribute by far the least number of observations (Table~\ref{tab:ico_platform_counts}). Figure~\ref{fig:data_split} illustrates the spatial distribution of observations, showing dense coastal coverage and sparser but widespread ship-based sampling across the open ocean.

\begin{table}[htbp]
 \centering
 \caption{Number of ICOADS observations by platform type (2015--2024).}
 \begin{tabular}{l r}
 \toprule
 \textbf{Platform type} & \textbf{Observations} \\
 \midrule
 Ship & 7\,929\,745 \\
 Drifting buoy & 179\,923 \\
 Moored buoy & 9\,677\,483 \\
 C-MAN station & 1\,518\,359 \\
 Coastal station & 3\,011\,731 \\
 Tide gauge & 8\,937\,940 \\
 \midrule
 \textbf{Total} & \textbf{34\,860\,848} \\
 \bottomrule
 \end{tabular}
 \label{tab:ico_platform_counts}
\end{table}

\subsection{ECMWF Reanalysis v5 (ERA5)}
\label{sec:era5}

ERA5 is the fifth-generation global climate reanalysis produced by the European Centre for Medium-Range Weather Forecasts (ECMWF) under the Copernicus Climate Change Service (C3S) \citep{hersbach2020era5,ECMWF,copernicus_climate_change_service}. It provides a consistent record of climate variables from January 1940 to the present. ERA5 offers hourly estimates of a wide range of atmospheric, land, and oceanic variables. The dataset covers the entire globe at a spatial resolution of $0.25^{\circ} \times 0.25^{\circ}$ and includes 137 vertical atmospheric levels, extending up to about 80~km altitude. In this work, we use ERA5 as a benchmark against which we compare our machine learning models. Since our focus is on predicting wind velocity and direction, we extracted the horizontal wind components ($U$ and $V$) at $10~\mathrm{m}$ height. To ensure consistency with in-situ records, ERA5 values were sampled at the time and location of each ICOADS observation. In each case, the closest available ERA5 time step is used as the reference value. Tests (not reported) showed that interpolation among the four nearest ERA5 grid points did not improve ERA5 accuracy relative to ICOADS observations, so we use the nearest-neighbor ERA5 value in this paper.

\subsection{Global Forecast System (GFS)}
\label{sec:gfs}

The Global Forecast System (GFS) is a global NWP model operated by NCEP \citep{ncep_gfs}. It provides gridded forecasts of atmospheric, oceanic, and land-surface variables. Forecasts extend up to 384 hours (16 days) ahead, with hourly lead times available for the first 120 hours and 3-hourly lead times thereafter. New forecast cycles are initialized four times per day (00, 06, 12, 18\,UTC) at a base horizontal resolution of $1^{\circ}$. GFS data are accessed through Google Earth Engine (GEE). While convenient, this interface restricts the set of available variables. We downloaded the $U$ and $V$ components of wind velocity at $10~\mathrm{m}$, air temperature at $2~\mathrm{m}$, specific humidity at $2~\mathrm{m}$, and relative humidity at $2~\mathrm{m}$.

GFS data are sampled at the closest available grid point in space and time for each ICOADS observation. However, unlike ERA5, which provides continuous hourly coverage, GFS is issued only four times daily (every 6 hours), each run producing forecasts with lead times up to 384 hours. To ensure a prediction horizon of 48 hours with the best available prediction from GFS, the 48 forecast values are distributed across eight consecutive cycles. As GFS cycles occur every 6 hours, we make sure to use the appropriate GFS value coming from the right cycle based on the lead time value and the observation measurement time. For an observation at 03 UTC, the 00 UTC GFS cycle provides the 1-, 2-, and 3-hour lead-time forecasts. We obtain the lead times from 4 to 9 hours from the preceding day's 18 UTC GFS cycle using the appropriate forecast hours. The process is illustrated in Figure~\ref{fig:gfs_schema} in the appendix.

\section{Methods: ORCA}
\label{sec:methods_model}

Predicting the future state of the atmosphere is inherently difficult and has been the focus of sustained research across both physics-based NWP and modern AI weather models. Instead of attempting to generate forecasts directly from observations, \citet{yang2024local_off_grid} reframed the problem by demonstrating that machine learning can systematically correct errors in existing NWP guidance. This correction-based perspective is attractive because it leverages the strengths of NWP while allowing ML models to focus on residual, data-driven improvements. Our work adopts this same approach.

The original forecasting task, using only past observed values, can be formulated as:
\begin{equation}
 \mathbf{w}(t + l\Delta t) = h\!\left( \mathbf{w}(t - b\Delta t : t) \right), 
 \label{eq:forecasting_eq}
\end{equation}
where $t$ denotes the current time, $\mathbf{w}$ represents a vector of weather observations, $l\Delta t$ is the lead time (the time into the future for the prediction), $b\Delta t$ is the historical time window used as input, and $h$ is a learned forecasting function that maps a window of past observations to a future wind state.

The alternative correction-based task is written as:
\begin{equation}
 \mathbf{w}(t + l\Delta t) = f(t + l\Delta t) \;+\; h_{\theta}\!\left( \mathbf{w}(t - b\Delta t : t),\; f(t - b\Delta t : t + l\Delta t) \right),
 \label{eq:correcting_eq}
\end{equation}
 where $f(t + l\Delta t)$ is the NWP forecast (GFS) at lead time $l\Delta t$ and $h_{\theta}$ is the learned correction network, parameterized by~$\theta$, that predicts a state-dependent additive adjustment conditioned on both past observations~$\mathbf{w}$ and NWP guidance~$f$. This notation distinguishes the direct forecasting function~$h$ in Eq.~\eqref{eq:forecasting_eq} from the correction model~$h_{\theta}$ in Eq.~\eqref{eq:correcting_eq}, and separates the NWP baseline~$f$ from the learned residual.

\subsection{General Framework} 
\label{sec:framework}
A major deviation from previous work \citep{yang2024local_off_grid} is that, in this work, we do not assume that past ocean observations remain at fixed locations across time steps. Instead, both the number and spatial distribution of available observations vary dynamically. The model must therefore predict at arbitrary positions, with a changing set of reference points between consecutive time steps. A central difficulty is the sparsity of observations over the Atlantic, which raises the question of which past points should be retained for prediction. 
To assess whether spatial or temporal proximity is more informative, we computed the Pearson correlation coefficients between target points and past observations under varying distance and time thresholds (Fig.~\ref{fig:correlation}). The results indicate that correlation is dominated by temporal proximity: the most recent measurements, even if geographically distant, are more predictive than older nearby observations. Consequently, the framework adopts a simpler approach: all ICOADS observations from the hour immediately preceding the target time are used as input, regardless of their spatial distance from the target. This design enables inference at arbitrary sets of positions for a specific time step across the ocean in a single pass, while maintaining predictive skill.

\subsection{Model Implementation} 

\begin{figure}[h]
 \centering
 \includegraphics[width=1\linewidth]{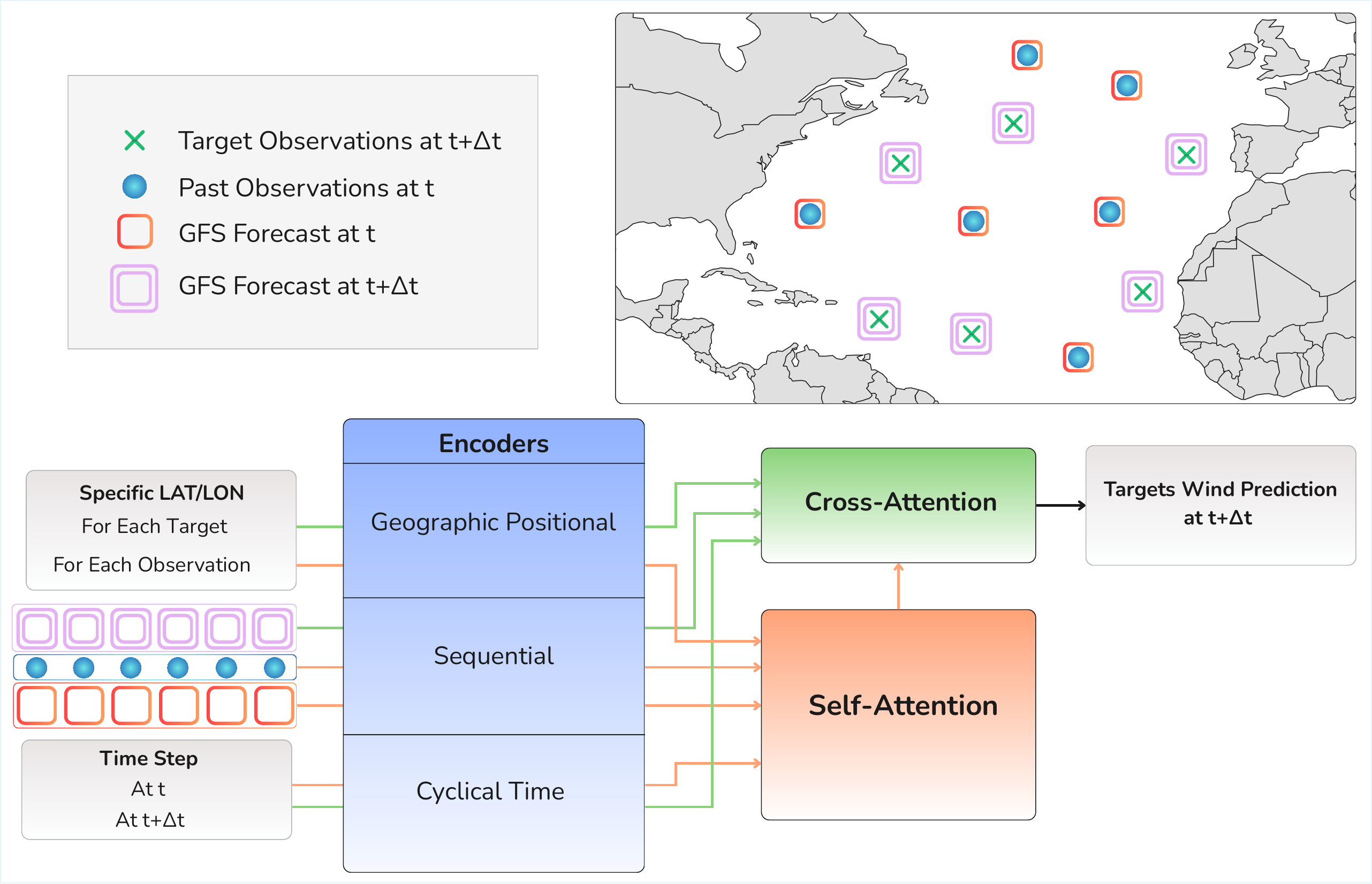}
 \caption{Schematic representation of the spatio--temporal learning framework. Each sample consists of in-situ observations at time t (blue circles), corresponding GFS forecasts at t (red squares) and t + $\Delta$t (magenta squares), and target observations at t + $\Delta$t (green crosses). The lower panels illustrate how each input is encoded through cyclical time, geographic positional, and sequential embeddings. Past observations and past GFS forecasts are processed through self-attention to capture temporal and spatial dependencies and learn the correction pattern, while cross-attention integrates information from GFS forecasts to predict target wind components at t + $\Delta$t.}

 \label{fig:model_implementation}
\end{figure}

 The model learns from pairs of past observations and historical forecast values coming from GFS at specific locations. The model must learn how to correct past GFS predictions based on the historical observations. It must also capture which pairs are more relevant to correct the GFS forecast for a given target point (Figure~\ref{fig:model_implementation}).

We use Vision Transformers (ViTs) due to their state-of-the-art performance and ability to encode positional information. Following \cite{kazemnejad2019positional}, positional embeddings $\mathbf{e}$ are added to the input embeddings $\mathbf{X}$, where each $\mathbf{x}_i$ represents the feature vector of the $i$-th pair element:
\begin{equation}
 \mathbf{X} = \left[ x_1 + e_1,\, x_2 + e_2,\, \ldots,\, x_n + e_n \right],
 \label{eq:pos_emb}
\end{equation}
This architecture is adopted as the basis of the model, with task-specific modifications.
The architecture follows an encoder--decoder design: the encoder applies self-attention over historical observation--GFS pairs to build a context representation, while the decoder uses cross-attention to condition predictions at new target locations on this context.

\subsubsection{Self-attention among historical observation-NWP forecast pairs} 

We process \{past observation, GFS forecast\} pairs through a self-attention mechanism (Figure~\ref{fig:model_implementation}), enabling the model to assign varying importance to each pair in the correction pattern. Let $\mathbf{X}$ be an input sequence of $n$ historical pairs $[x_1, x_2, \ldots, x_n] \in \mathbb{R}^{n \times d}$, where each $x_i$ concatenates the observed state and the corresponding GFS forecast at a given time. Self-attention \citep{raschka2023selfattention} produces a new representation as: 
\begin{equation}
 Attention(\mathbf{Q}, \mathbf{K}, \mathbf{V}) = \mathrm{softmax}\!\left(\frac{\mathbf{QK}^\top}{\sqrt{d_k}}\right)\mathbf{V},
 \label{eq:self-attention}
\end{equation}
where $\mathbf{Q}, \mathbf{K}, \mathbf{V} \in \mathbb{R}^{n \times d_k}$ are the query, key, and value matrices obtained from linear projections of $\mathbf{X}$, and $d_k$ is the dimension of the projected features. The dot product between queries and keys quantifies similarity between the pairs. 

In practice, Multi-Head Self-Attention (MHSA) \citep{vaswani2017attention} is applied to this sequence to capture interactions between all pairs across multiple representation subspaces: 
\begin{equation}
 MHSA(\mathbf{X}) = \big[\, head_1 \,\|\, \ldots \,\|\, head_h \,\big]\mathbf{W}^O,
 \label{eq:multi_at}
\end{equation}
where each $head_i = Attention(\mathbf{X}\mathbf{W}^Q_i, \mathbf{X}\mathbf{W}^K_i, \mathbf{X}\mathbf{W}^V_i)$ and $\|$ denotes concatenation. 

 Through the attention scores, the model learns which historical observation-GFS forecast pairs are most relevant for correcting the GFS forecast at the target points, enabling flexible weighting of past data in generating the corrected prediction.

\subsubsection{Cross-attention between past observations and target predictions} 
The second key component of the model is cross-attention, which enables the network to identify which past observations are most relevant for each target prediction. Whereas self-attention allows tokens within the same sequence to attend to one another, cross-attention relates tokens from two different sequences \citep{lin2021cat}.

The attention module receives two inputs:
\begin{itemize}
 \item a \textbf{query} sequence $\mathbf{Q} \in \mathbb{R}^{n_q \times d_k}$ (e.g., target tokens),
 \item and a \textbf{key}-\textbf{value} sequence $\mathbf{K} \in \mathbb{R}^{n_k \times d_k}$, $\mathbf{V} \in \mathbb{R}^{n_k \times d_v}$ (e.g., past observation--forecast pairs).
\end{itemize}

The cross-attention operation is formally defined as:
\begin{equation}
 CrossAttention(\mathbf{Q}, \mathbf{K}, \mathbf{V}) 
 = \mathrm{softmax}\!\left(\frac{\mathbf{QK}^\top}{\sqrt{d_k}}\right)\mathbf{V},
 \label{eq:cross_attention}
\end{equation}
where $n_q$ and $n_k$ denote the number of query and key/value tokens, respectively, $d_k$ the dimension of queries and keys, and $d_v$ the dimension of the values. This formulation allows each query token to attend selectively to all tokens in the key-value sequence, thereby extracting the most relevant contextual information. 

As with self-attention, cross-attention is implemented using multiple heads: 
\begin{equation}
 MHA_{\mathrm{cross}}(\mathbf{Q}, \mathbf{K}, \mathbf{V}) 
 = \big[\, head_1 \,\|\, \ldots \,\|\, head_h \,\big]\mathbf{W}^O,
 \label{eq:mha_cross}
\end{equation}
where each $head_i = Attention(\mathbf{QW}^Q_i, \mathbf{KW}^K_i, \mathbf{VW}^V_i)$ and $\|$ denotes concatenation. 

Cross-attention layers are stacked after self-attention within the decoder blocks.

This mechanism is particularly important in our setting, where the available observations vary over time and space, and the model must dynamically determine which inputs are most informative for each prediction.

\subsubsection{Masking for Irregular Observations} 
A distinctive feature of our setting is that both the number and spatial distribution of observations vary at every time step. To ensure the transformer operates consistently under these conditions, we construct binary masks that specify which tokens are valid inputs and which target points should contribute to the loss. These masks are applied during both self-attention and cross-attention layers, preventing the model from attending to nonexistent observations, and during loss computation, ensuring that errors are only accumulated over available targets. This design allows the architecture to handle changing observation sets dynamically without requiring imputation or fixed spatial grids.

\subsubsection{Position and Time Embedding} 
\label{sec:pos_time_emb}
The positional embeddings~$\mathbf{e}$ in Eq.~\eqref{eq:pos_emb} are the sum of two components detailed below: a cyclical time encoding capturing daily and annual periodicities, and a geographic positional encoding based on spherical harmonics. These correspond to the ``Cyclical Time'' and ``Geographic Positional'' blocks in Figure~\ref{fig:model_implementation}, respectively; both are added to the input features before entering the self-attention and cross-attention layers.
To fully exploit the capabilities of the ViT architecture, we add both temporal and positional embeddings to the feature representations before entering the transformer layers. We encode each timestamp using four cyclical features capturing daily and annual periodicities: $\sin\!\left(\tfrac{2\pi d}{366}\right)$, $\cos\!\left(\tfrac{2\pi d}{366}\right)$, $\sin\!\left(\tfrac{2\pi h}{24}\right)$, and $\cos\!\left(\tfrac{2\pi h}{24}\right)$ where $d$ is the day of the year and $h$ the hour of the day. This encoding preserves the cyclic nature of time and provides a compact representation for downstream learning, and has been widely adopted in transformer-based forecasting models \citep{Pospichal2022Solar, Su2025systematic}.

We adopt the geographic positional encoding scheme of~\citep{russwurm2023locationencoder}. This approach combines spherical harmonic basis functions with a Sinusoidal Representation Network (SirenNet). Concretely, each latitude-longitude coordinate is first projected into spherical harmonic coefficients (up to a chosen degree), which are then processed by a neural network to produce a learned embedding vector. This hybrid encoding captures both global spherical geometry (avoiding distortions at high latitudes or across the date line) and fine-grained local structure, making it well-suited for geospatial modeling in our project.

\section{Experimental Setup}
\label{sec:experimental}

\subsection{Forecasting Task and Baselines} 
The forecasting task is to predict 10-m wind velocity at arbitrary offshore locations up to 48~h, using as input the most recent set of observations combined with the GFS forecast field. Predictions are evaluated against ICOADS observations (Section~\ref{sec:icoads}). We compare ORCA against three baselines: the Global Forecast System (GFS), which provides the starting point for correction; ERA5 reanalysis, used as a reference benchmark; and a MOS-style linear regression \citep{glahn1972mos} that predicts corrected wind components from GFS forecasts and spatially aggregated observations (see \ref{sec:lin_reg_baseline} for details). We report comparisons across multiple lead times and stratify by platform type and spatial region.

\subsection{Data Splitting} 
To ensure generalization to unseen conditions, we split the dataset chronologically: the first 80\% of data temporally are used for training (27.8M samples), the next 10\% for validation (3.6M), and the final 10\% for testing (3.4M) (Fig.~\ref{fig:set_split}). This temporal split prevents data leakage and ensures the model is evaluated on future time periods not seen during training.
Within each temporal split, the available observation locations vary dynamically at each time step based on which platform provided measurements at that time. Consequently, the target prediction points--- i.e. all the observations at some future lead time---also change over time steps. 

This design ensures the model learns to predict wind corrections at arbitrary ocean locations, promoting spatial generalization and enabling robust performance at previously unseen geographic positions during validation and testing.

\subsection{Model Training}

We train the models to minimize the mean vector wind error between predicted and observed 10-m wind vectors,
\begin{equation}
 \mathcal{L} = \frac{1}{N}\sum_{i=1}^{N} \lVert \hat{\mathbf{w}}_i - \mathbf{w}_i \rVert_2
 = \frac{1}{N}\sum_{i=1}^{N} \sqrt{(\hat{u}_i - u_i)^2 + (\hat{v}_i - v_i)^2}\,,
 \label{eq:loss}
\end{equation}
where $\hat{\mathbf{w}}_i = (\hat{u}_i, \hat{v}_i)$ and $\mathbf{w}_i = (u_i, v_i)$ are the predicted and observed wind vectors at valid target location~$i$, and the sum runs only over valid positions. This metric simultaneously penalizes errors in both wind speed and direction, and is used consistently for training, early stopping, and evaluation throughout this work. We perform optimization using AdamW with weight decay, an initial learning rate of $1 \times 10^{-4}$, and cosine learning-rate scheduling with warm restarts. We trained our models for 100 epochs with early stopping, based on validation mean vector wind error and a patience of 25 epochs (i.e., training stops if no improvement is observed for 25 consecutive epochs). Our experiments are run on NVIDIA Tesla V100 GPUs (16 GB). Training a single model requires approximately 24 hours. To ensure reproducibility, all runs use fixed random seeds for data shuffling, parameter initialization, and dropout.

\subsection{Hyperparameters and Architecture}
The transformer encoder consists of 8 layers with 8 attention heads each, a hidden dimension ($d_k$) of 128. Temporal embeddings capture both hourly and annual cycles, while spatial embeddings use spherical harmonics combined with a sinusoidal representation network (Section~\ref{sec:pos_time_emb}). Cross-attention layers condition target points on the encoded set of observation-forecast pairs, enabling inference at arbitrary spatial coordinates.

\section{Results} 
\label{sec:results}

We present our results in three parts. First, we evaluate the global performance of ORCA, analysing its performance throughout the different platform types and in comparison with the global numerical forecasting models. Second, we conduct an extensive spatial analysis to identify potential regional patterns and areas where the model performs well or struggles. Finally, we perform ocean-wide inference, leveraging the model's capability to generate predictions at arbitrary locations, across the entire ocean domain. We compare the gridded model outputs against the full GFS field.

We focus on specific oceanic case studies to examine in detail how the wind field is modified in localized regions, and how the spatial density of nearby observations influences these corrections.

\subsection{Global Performance} 
\begin{figure}[h]
 \centering
 \begin{subfigure}[t]{0.52\linewidth}
 \centering
 \includegraphics[width=\linewidth,height=6cm,keepaspectratio]{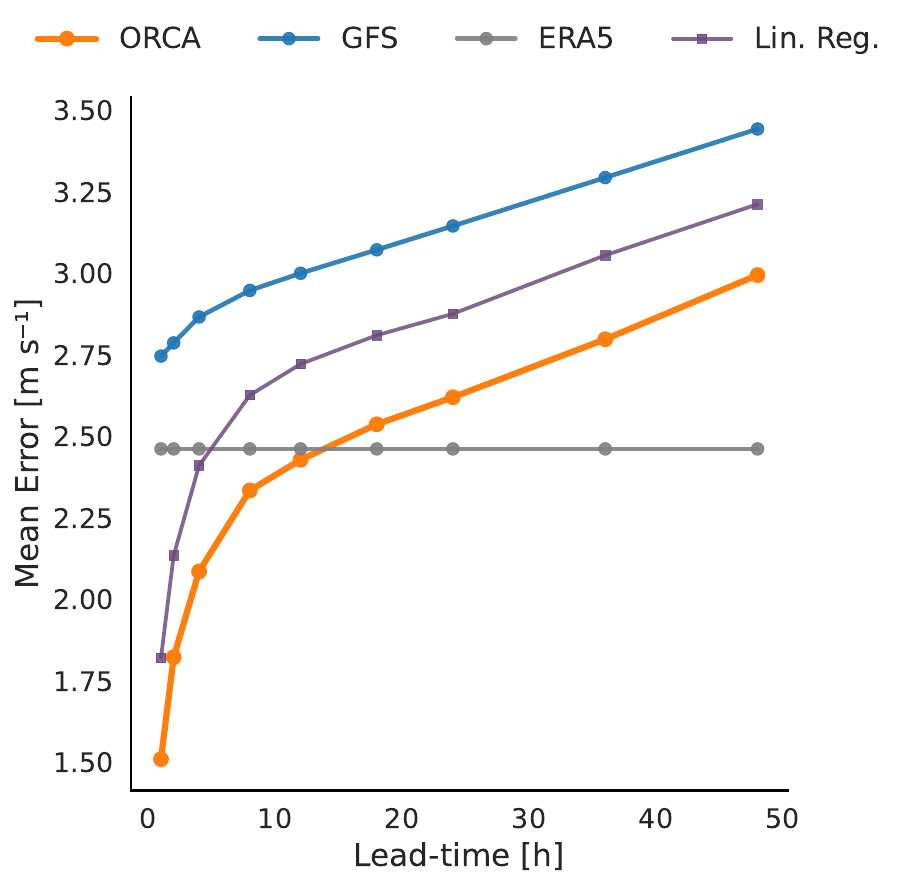}
 \par\vspace{0pt}
 \caption{Mean vector wind error of wind predictions as a function of forecast lead time. ORCA (orange) is compared against the linear regression baseline (purple), GFS forecasts (blue), and ERA5 reanalysis (grey).}
 \label{fig:general_results}
 \end{subfigure}
 \hfill
 \begin{subfigure}[t]{0.44\linewidth}
 \centering
 \includegraphics[width=\linewidth,height=6cm,keepaspectratio]{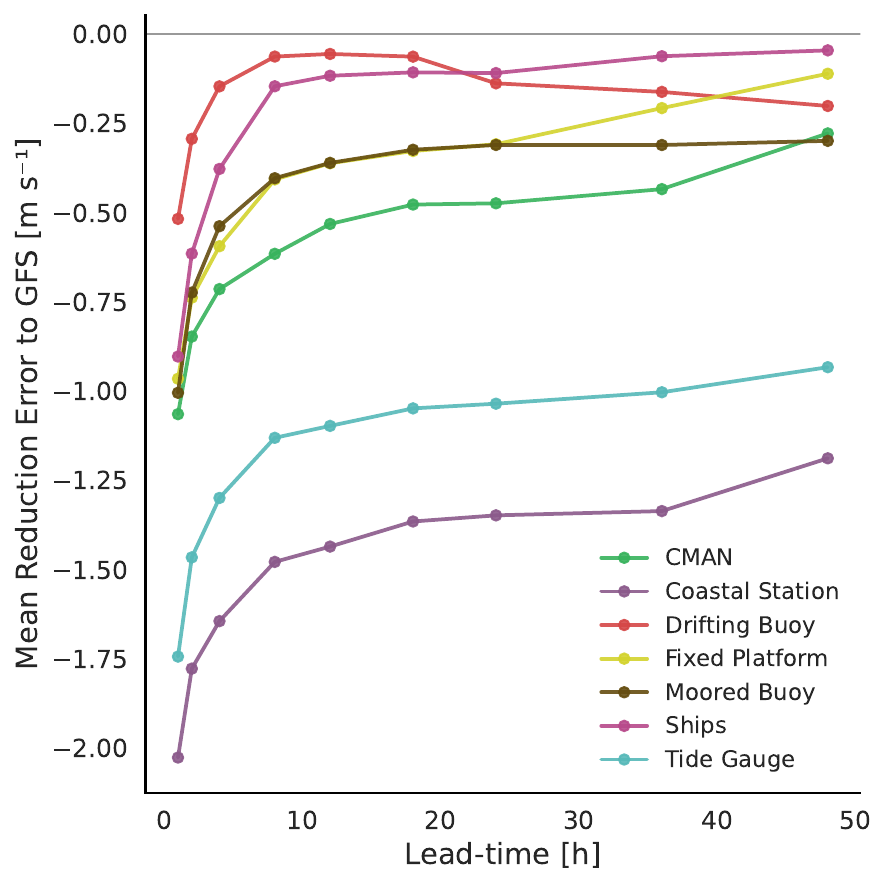}
 \par\vspace{0pt}
 \caption{Mean vector wind error difference between ORCA and GFS forecasts, grouped by observation platform type. Negative values indicate an error reduction relative to GFS.}
 \label{fig:pt_results}
 \end{subfigure}
 \caption{Comparison of model performance across forecast lead times and observation platforms. 
(a) Overall accuracy relative to the linear regression baseline, GFS, and ERA5. (b) Error reduction by platform type.}
 \label{fig:general_and_pt_results}
\end{figure}

ORCA consistently improves upon GFS forecasts for all lead times up to 48~h. 
Figure~\ref{fig:general_results} compares the performance of ORCA, the linear 
regression baseline, GFS, and ERA5 across lead times, while 
Table~\ref{tab:rmse_leadtimes} quantifies the improvements. ORCA reduces GFS 
error by 45\% at 1~h lead time (from 2.75 to 1.51~m\,s$^{-1}$) and by 13\% 
at 48~h (from 3.44 to 2.99~m\,s$^{-1}$), outperforming ERA5 up to 12~h. The 
linear regression baseline also improves over raw GFS (by 6.7--33.8\%), but 
ORCA outperforms it by an additional 7--17\% at all lead times.

\begin{table}[htbp]
 \centering
 \footnotesize
 \caption{Mean vector wind error (m\,s$^{-1}$) of 10-m wind for several lead times, 
          and relative improvement of ORCA and the linear regression baseline with respect to GFS.}
 \begin{tabular}{c S[table-format=1.2] S[table-format=1.2] S[table-format=1.2] S[table-format=1.2] S[table-format=2.1] S[table-format=2.1]}
 \toprule
 \textbf{Lead time} & 
 {\textbf{ORCA}} & 
 {\textbf{Lin.~Reg.}} & 
 {\textbf{GFS}} & 
 {\textbf{ERA5}} & 
 {\textbf{ORCA\textsuperscript{a}}} &
 {\textbf{Lin.~Reg.\textsuperscript{a}}} \\
 (h) & {(m\,s$^{-1}$)} & {(m\,s$^{-1}$)} & {(m\,s$^{-1}$)} & {(m\,s$^{-1}$)} & {(\%)} & {(\%)} \\
 \midrule
  1 & 1.51 & 1.82 & 2.75 & 2.46 & 45.1 & 33.8 \\
  2 & 1.82 & 2.14 & 2.79 & 2.46 & 34.6 & 23.5 \\
  4 & 2.09 & 2.41 & 2.87 & 2.46 & 27.3 & 16.0 \\
  8 & 2.33 & 2.63 & 2.95 & 2.46 & 20.9 & 10.9 \\
 12 & 2.43 & 2.72 & 3.00 & 2.46 & 19.1 &  9.3 \\
 18 & 2.54 & 2.81 & 3.07 & 2.46 & 17.5 &  8.6 \\
 24 & 2.62 & 2.88 & 3.15 & 2.46 & 16.8 &  8.6 \\
 36 & 2.80 & 3.06 & 3.29 & 2.46 & 15.1 &  7.2 \\
 48 & 2.99 & 3.21 & 3.44 & 2.46 & 13.1 &  6.7 \\
 \bottomrule
 \end{tabular}
 
 \label{tab:rmse_leadtimes}
 \vspace{0.5em}
 \raggedright\footnotesize
 \textsuperscript{a}\,Improvement = $(\text{GFS} - \text{Model}) / \text{GFS} \times 100\%$.
\end{table}

We tested several architectural variations to assess potential improvements beyond the baseline configuration. These changes are summarized in Table~\ref {tab:ablation_study}. None of the modifications, including the use of longer temporal context, additional GFS predictors, platform metadata, or residual GFS connections, yielded measurable gains. 

The following analysis uses a single model trained on all platform types jointly; results are stratified by platform at evaluation time.
To further evaluate model behavior, we stratify performance by platform type and by individual observation station. This assessment aims to identify whether certain observation platforms pose greater forecasting challenges or whether specific stations consistently yield higher prediction errors. Figure~\ref{fig:pt_results} presents the mean error reduction relative to GFS as a function of lead time and platform category. On average, all platform types exhibit improved skill compared to GFS, though the magnitude of improvement varies. The largest gains are observed for coastal stations, followed by tide gauges, CMAN stations, fixed platforms, moored buoys, and ships, while drifting buoys show the smallest improvement.
Notably, because the dataset is split temporally, ships and drifting buoys are evaluated at locations that differ from those seen during training---ships follow different routes and drifting buoys move significant distances between the training and test periods. The requirement that the model generalize to previously unseen locations is one likely explanation for the relatively smaller improvement on ship and drifting buoy observations.

To evaluate model performance under high-wind conditions, we stratified results by observed wind speed using the 90th and 95th percentiles (10.9 and 12.9~m\,s$^{-1}$, respectively) as thresholds (Figure~\ref{fig:extreme_wind}). At short lead times (1--8~h), the model improves over GFS across all wind regimes, including the top 5\% of wind speeds. However, the relative improvement diminishes more rapidly with lead time for extreme winds: for the top 5\%, gains fall below zero near 18~h, reaching approximately $-$5 to $-$7\% at 36--48~h. In other words, at longer lead times, GFS is more accurate for high-wind forecasting than ORCA.

\subsection{Spatial Performance}
\begin{figure}[h]
 \centering
 \includegraphics[width=1\linewidth]{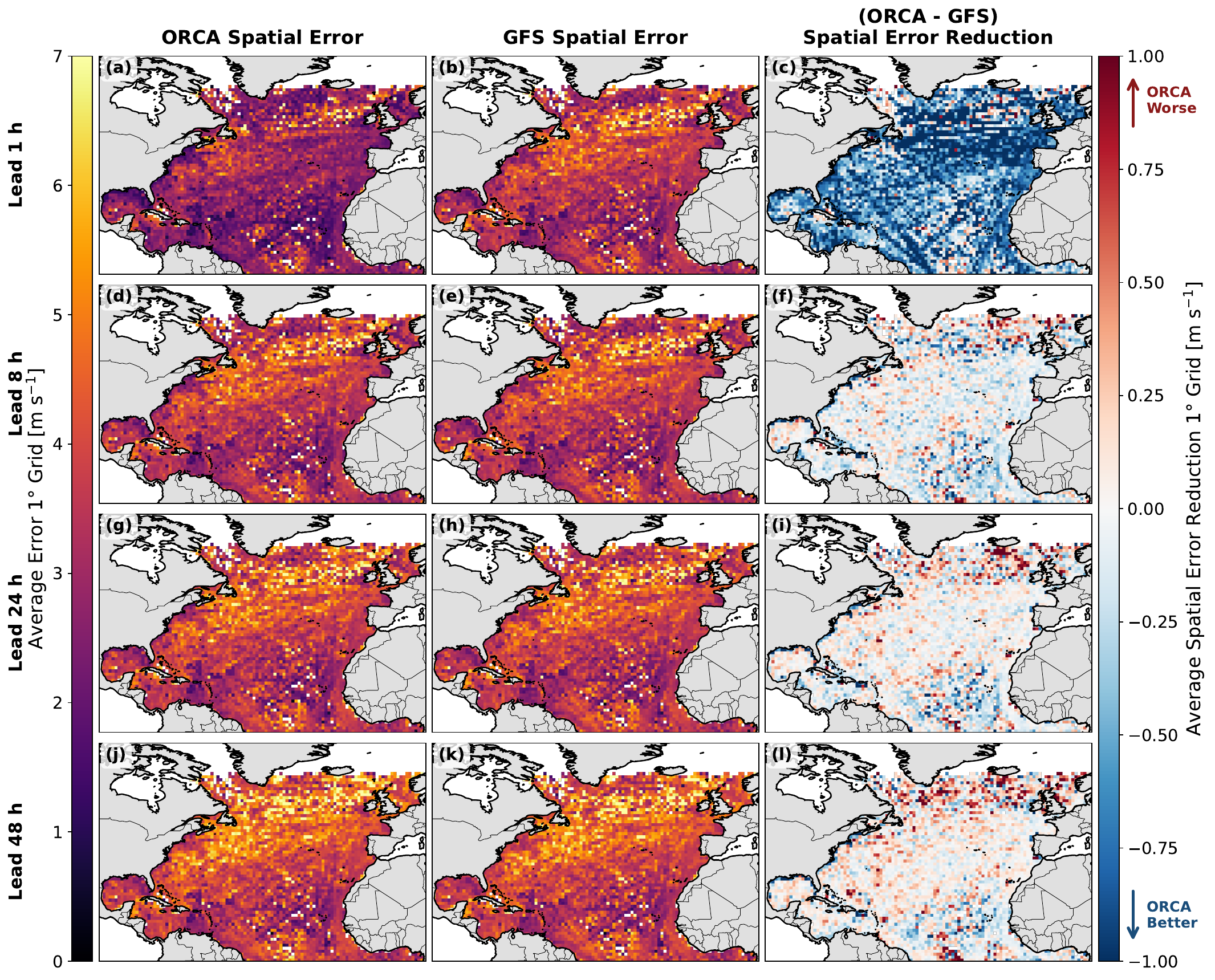}
 \caption{Spatial distribution of wind-speed prediction errors for ORCA and the GFS baseline at lead times of 1 h, 8h, 24 h, and 48 h. In the right column, blue indicates ORCA outperforms GFS and red indicates GFS outperforms ORCA.}
 \label{fig:spatial_error}
\end{figure}

To assess how corrections vary across regions and lead times, Figure~\ref{fig:spatial_error} compares spatial distributions of average wind-speed errors for the proposed model and the GFS baseline at 1, 8, 24, and 48~hour lead times. The left and middle columns show absolute errors for the machine-learning corrector and GFS, respectively, while the right column illustrates their difference (model~--~GFS), where blue indicates improvement and red degradation. Across all lead times, the model produces lower errors over most of the Atlantic domain, with widespread blue shading in the difference panels. The strongest improvements are found along the eastern seaboard of North America, the Caribbean, and western European coasts, as well as along major shipping corridors. In these regions, local biases in GFS are systematically reduced by the observation-informed corrections. As forecast horizon increases, both models exhibit higher errors. Nevertheless, the corrector maintains clear gains up to 48~hours, with the difference maps remaining predominantly blue across the basin. The relative advantage gradually diminishes in magnitude and spatial extent, especially over the central and northern Atlantic, where observational coverage is sparse. Small, isolated regions of degradation appear at high latitudes. In general, ORCA performs better at locations with more previous observations available nearby (Figure \ref{fig:density_improvement}). However, we also observe at low latitudes, where data is sparse, that ORCA consistently improves the GFS original predictions. 

Model errors inherit the spatial structure of GFS errors, as GFS serves as the baseline for the machine-learning corrector. Error varies geographically across the basin, with higher values at high latitudes and lower values at low latitudes. Mid- and high latitudes typically have more energetic and variable winds that are harder to predict accurately \citep{Zhang2024_wind_climatology}. 
Notably, the model's performance relative to GFS is spatially heterogeneous. In regions where GFS exhibits systematic biases, the corrector effectively reduces errors. However, for some areas---particularly at higher lead times---the model shows limited improvement or occasional degradation.

\subsection{Global Inference} 

\begin{figure}[h]
 \centering
 \includegraphics[width=1\linewidth]{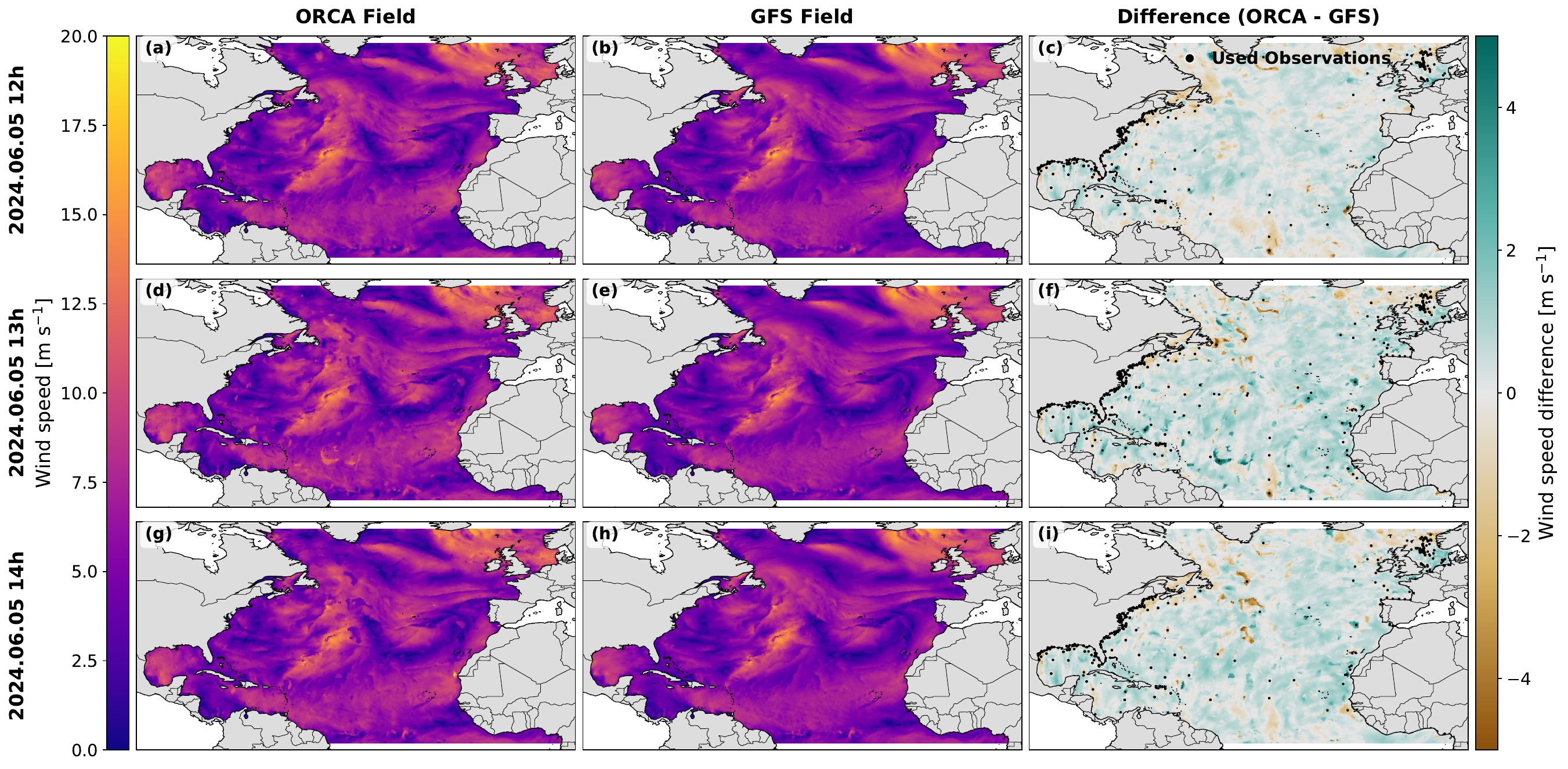}
 \caption{Comparison of ORCA wind fields, GFS field, and their differences across three consecutive time steps at 1h lead-time.}
 \label{fig:grid}
\end{figure}

To evaluate the model predictions across the entire Atlantic, we performed global inference on a regular grid of points (0.25$^\circ$ resolution) at multiple time steps. 

The results reveal clear differences between the corrected fields and the original GFS forecasts (Figure \ref{fig:grid}). In particular, ORCA introduces local adjustments in the vicinity of past observations, where it modifies the predicted wind velocities for subsequent time steps. This results in fields that are less spatially smooth than GFS, reflecting the localized influence of recent measurements. This is most obviously visualized in the difference field between our forecast and GFS (Figure \ref{fig:grid}, right column).

\begin{figure}[h]
 \centering
 \includegraphics[width=1\linewidth]{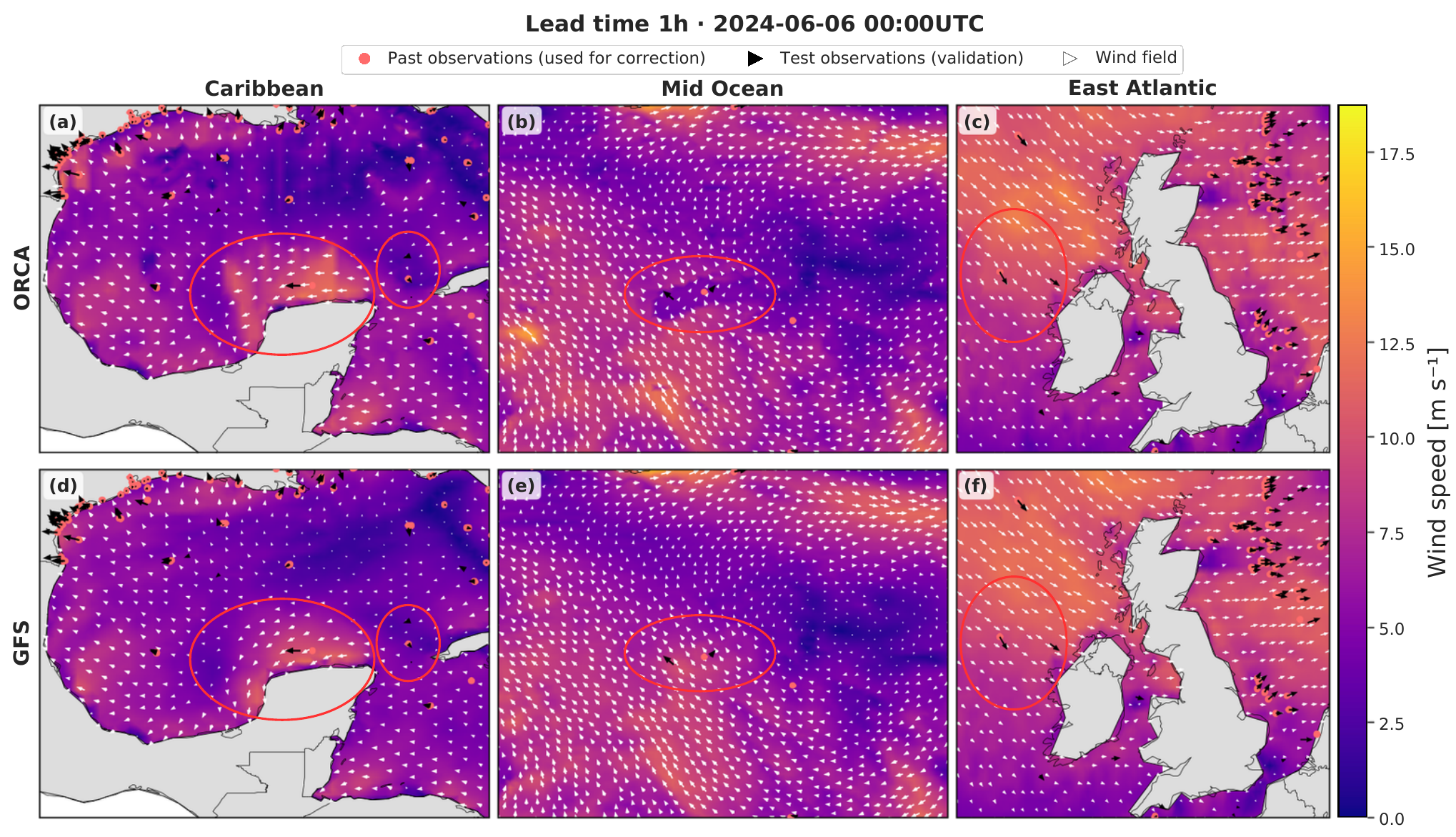}
 \caption{Case studies illustrating localized wind-field corrections across three representative regions.}
 \label{fig:case_studies}
\end{figure}

 In some cases, the model also generates isolated adjustments in regions without direct observational input. This likely arises because the learned location encoders and cross-attention weights do not enforce strict spatial locality: a distant observation can produce a small but nonzero attention weight at a far-away target, resulting in minor corrections even where no nearby measurements exist. Whether these adjustments are physically meaningful or artifacts of extrapolation is difficult to determine in general; the case studies below illustrate both situations. We highlight several case studies to illustrate how the model adjusts the wind field over the ocean, using past observations to better align with subsequent measurements (Figure~\ref{fig:case_studies}). In Figures~\ref{fig:case_studies}(c) and~\ref{fig:case_studies}(f), the corrected field northwest of Ireland is adjusted toward a pattern that more closely matches the observed values. Similarly, in Figures~\ref{fig:case_studies}(a) and~\ref{fig:case_studies}(d), a distinct circular region of adjusted wind speed is visible around a prior observation northwest of Cuba, marked by the red circle, where the model refines both wind direction and intensity relative to the GFS baseline.

\section{Discussion}

\textbf{Our results show that data-driven methods can meaningfully correct NWP and deliver substantial improvements.} The transformer-based corrector consistently reduces surface-wind errors across all lead times, outperforming GFS by 45\% on average at 1~h and maintaining a 13\% average improvement at 48~h. These gains demonstrate that in-situ observations contain exploitable information about short-term wind patterns and that a learned corrector can translate these data into improved predictions. GFS provides a starting point that is physically grounded but biased, while the transformer focuses on learning local patterns and state-dependent errors from the latest observations through its attention mechanisms. Our early experiments over the ocean and prior work have both confirmed that the transformer architecture is better suited to capture observation-conditioned error patterns than simpler models like MLPs, LSTMs, and linear regression \citep{yang2024local_off_grid} (see Appendix Figure~\ref{fig:baseline_comparison}, Table~\ref{tab:baseline_comparison}, and Table~\ref{tab:rmse_leadtimes} for a direct comparison). ORCA also outperforms ERA5 up to 12~h, indicating that short-term, observation-driven corrections can surpass even retrospectively optimized reanalysis fields. 

However, these aggregate gains mask important differences across wind-speed regimes (Figure \ref{fig:extreme_wind}). Stratifying by observed wind speed reveals that the model improves over GFS at all intensities for short lead times (1--8~h), including the strongest 5\% of winds ($\geq$12.9~m\,s$^{-1}$). At longer horizons, the advantage narrows for high-wind events and reverses beyond approximately 18~h for the top 5\%, reaching $-$5 to $-$7\% degradation at 36--48~h. This likely reflects both the rarity of extreme observations in the training data---the top 5\% accounts for fewer than 2M of 34.8M samples---and the faster decorrelation of intense, transient weather patterns. Improving performance under extreme conditions through event-weighted training or oversampling remains an important direction for future work.

\noindent\textbf{Model performance improves with higher observational density in space and time.}
The most consistent error reductions compared to GFS appear where recent observations are available, in particular along coasts and major shipping routes, while gains diminish in open-ocean regions where measurements are sparser. In data-sparse areas, ORCA's corrections naturally weaken, and predictions revert toward the GFS baseline. However, performance can also degrade below GFS skill when the model overfits to isolated measurements without learning consistent local correction patterns, as suggested by the scatter in low-density grid cells in Figure~\ref{fig:density_improvement}. 

Performance improvements also vary systematically across platform types. Stationary platforms---including coastal stations, tide gauges, C-MAN stations, and fixed offshore platforms---exhibit greater improvements than moving platforms such as drifting buoys and ships. This hierarchy likely reflects the model's ability to learn tailored, location-specific corrections for fixed platforms, exploiting persistent error structures similar to fixed-site correction systems studied in \citet{yang2024local_off_grid}. In contrast, moving platforms present a harder generalization problem: ships follow dynamic routes that may traverse poorly sampled regions, while drifting buoys can enter ocean areas with few training observations. Drifting buoys may represent the hardest correction challenge because, unlike ships that tend to follow established corridors, they drift unpredictably into sparsely observed ocean regions.
It is important to distinguish the architectural capability of arbitrary-location inference from empirical evidence of spatial generalization. ORCA can predict at any ocean coordinate by design, using learned location encoders that are not tied to a fixed station set. For mobile platforms (ships and drifting buoys), the temporal data split ensures that test-time locations genuinely differ from training, providing direct evidence of spatial generalization. For stationary platforms, the improvements instead reflect location-specific learned corrections at sites seen during training. A wider testing design---withholding entire regions or platform types from training---remains an important direction for future work.

These platform-dependent differences reflect a broader structural challenge in the marine observing system. Although ICOADS provides the most comprehensive historical archive of marine meteorological observations, its spatial coverage remains biased toward shipping corridors and coastal regions \citep{climatedataguide_icoads}, leaving large oceanic areas persistently undersampled. The long-term decline in voluntary observing ships further threatens the continuity of in-situ data streams essential for machine-learning-based corrections \citep{awe_icoads_decline}. Emerging opportunities, such as integrating meteorological reports transmitted through the global Automatic Identification System (AIS) network, may partially mitigate this gap by supplying near-real-time weather measurements from thousands of vessels, potentially densifying coverage in historically sparse regions and improving correction performance \citep{erdc_ais_weather, researchgate_ais_weather}.

\noindent\textbf{Operational feasibility is demonstrated at basin scale with modest computational requirements.}
Our global inference experiment indicates that the corrector can be applied across a 0.25$^\circ$ Atlantic grid with modest computational cost, making basin-scale updates feasible on operational timescales. Inference consists of a single forward pass evaluated at arbitrary target coordinates using a single NVIDIA Tesla V100 GPU (16 GB), generating corrected fields in under 5 minutes. This enables rapid regeneration of forecasts as new measurements arrive, integrating into existing pipelines that ingest GFS forecasts and available in-situ observations. While the system is already suitable for operational deployment, there remain many opportunities for future work. Beyond deterministic forecasts, operational products may also benefit from introducing uncertainty quantification through probabilistic or conformal approaches to support ensemble and risk-aware decision-making in operational contexts. Likewise, while this study focuses on the Atlantic basin, extending or retraining the system for other ocean basins will test its robustness to differing climatologies and observational geometries.

\section{Conclusion}
\label{sec:conclusions}

We presented ORCA (Observation-informed Real-time Correction with Attention), a transformer-based correction system for marine wind forecasts. ORCA merges heterogeneous in-situ measurements with global forecasts across the Atlantic Ocean and can be evaluated at arbitrary points in the study region, avoiding the need for predefined grids. Using GFS as a physical prior, ORCA reduces forecast errors by 45\% at 1-hour lead time and 13\% at 48 hours, with improvements observed across ships, buoys, and coastal stations. 
Rather than replacing NWP, ORCA serves as a correction mechanism that adjusts large-scale forecasts using the latest available observations. Its performance gains and computational efficiency highlight the value of combining physical priors with learned, data-driven adjustments, particularly in regions where in-situ data is abundant. The spatial variability of improvements and the dependence on observational density emphasize the continued importance of robust marine observing networks.
Ultimately, our work aims to strengthen machine-learning-assisted marine forecasting for applications ranging from offshore renewable energy operations to maritime safety and navigation.

\section*{Code and Data Availability}
The code and data for this paper are available on GitHub (https://github.com/Earth-Intelligence-Lab/marine-wind-forecasting) and Zenodo (https://zenodo.org/records/17793234).

\section*{Conflict of Interest declaration}
The authors declare there are no conflicts of interest for this manuscript.

\acknowledgments
This work was supported by the MIT Maritime Consortium and the MIT-Sea Grant Doherty Professorship.

\bibliography{agusample}

\newpage

\appendix

\section{Appendix}

\begin{figure}[h]
 \centering
 \includegraphics[width=1\linewidth]{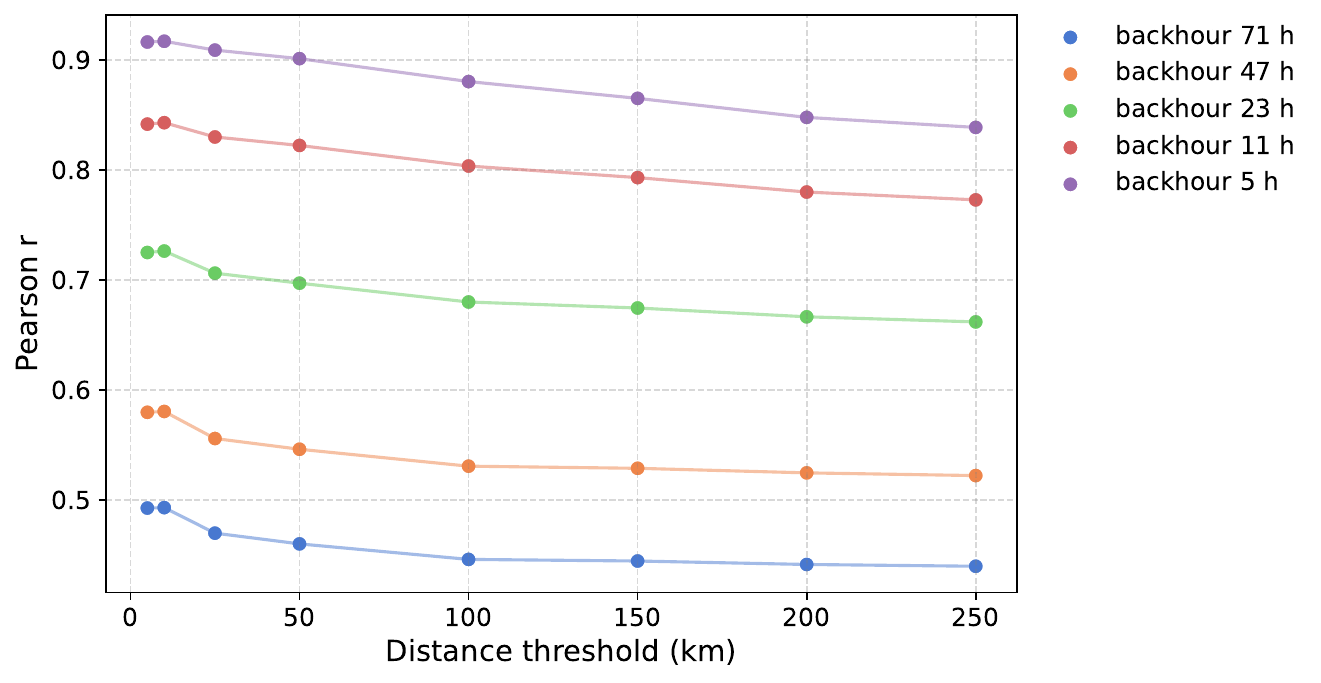}
 \caption{Pearson correlation ($r$) between target wind observations and past measurements as a function of spatial and temporal separation. 
 The x-axis shows the maximum spatial distance threshold (in kilometers) within which past observations are considered, and the colored lines correspond to different temporal windows ($\Delta t_{\max}$) ranging from 5 to 71~h, as indicated in the legend. 
 Each curve therefore represents how correlation decreases with increasing spatial distance for a fixed temporal window. 
 The results are computed using pairs of in-situ observations separated by approximately one hour ($\min\Delta t = -1$~h). 
 Correlations generally decrease both with distance and with increasing temporal window size, indicating that temporal proximity has a stronger influence on correlation than spatial distance.
 }
 \label{fig:correlation}
\end{figure}

\begin{table}[htbp]
 \centering
 \begin{tabular}{l c}
 \toprule
 \textbf{Modification} & \textbf{ $\Delta$ error} \\
 \midrule
 Two previous hours of observations & $<$\,0.01~m\,s$^{-1}$ \\
 Additional GFS variables (Temperature, Humidity, $q$) & $<$\,0.01~m\,s$^{-1}$ \\
 Platform/object type encoding & $<$\,0.01~m\,s$^{-1}$ \\
 Residual GFS connection after cross-attention & $<$\,0.01~m\,s$^{-1}$ \\
 \bottomrule
 \end{tabular}
 \caption{Summary of architectural variations tested. All differences in mean vector wind error were below 0.01~m\,s$^{-1}$ at all lead times, indicating no meaningful improvement over the baseline configuration.}
 \label{tab:ablation_study}
\end{table}

\begin{figure}[h]
 \centering
 \includegraphics[width=1\linewidth]{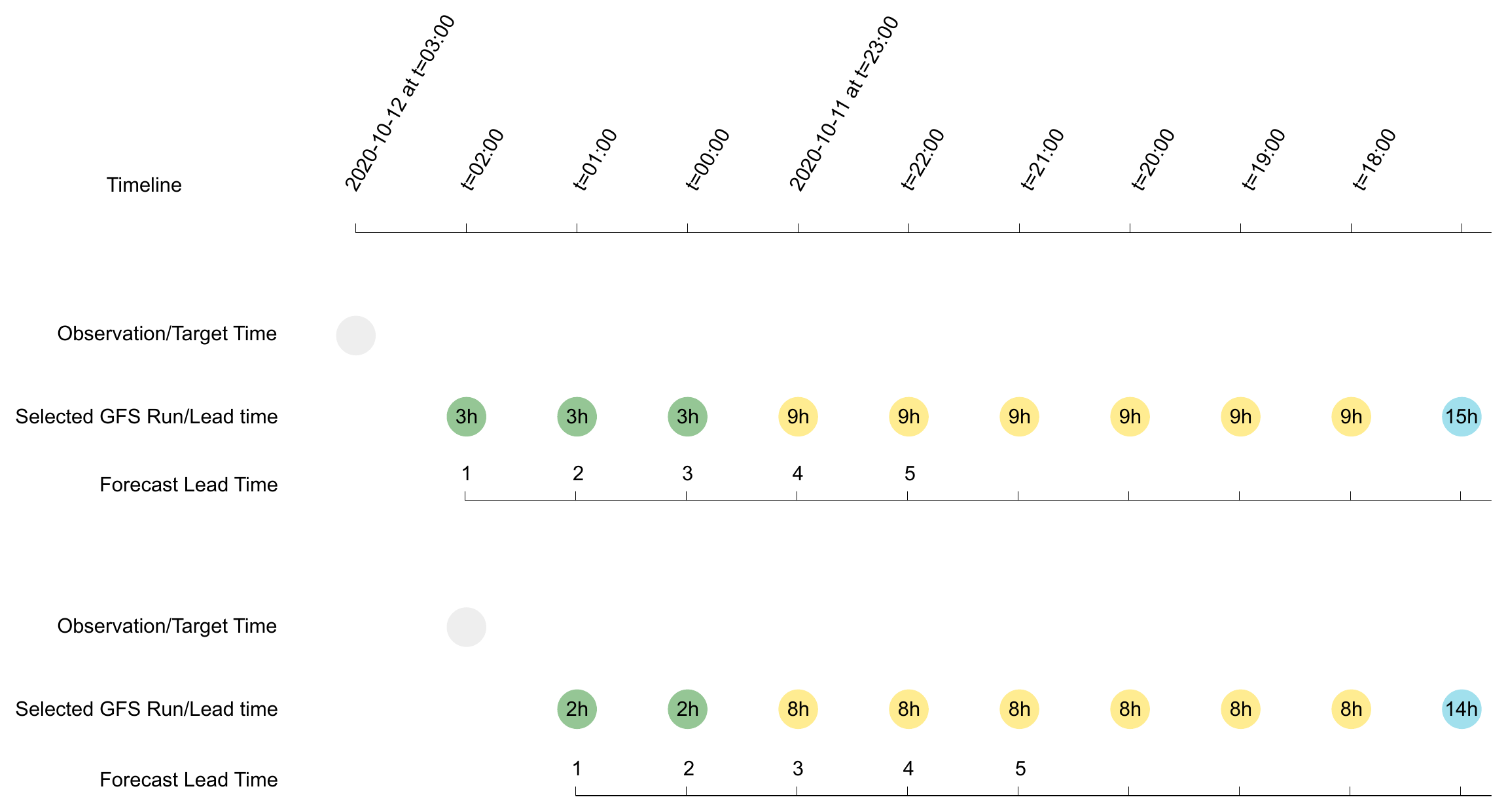}
 \caption{
 Schematic illustration of the GFS forecast cycle and its alignment with observation times. 
 Each GFS forecast is initialized every 6~hours and provides predictions at multiple lead times. 
 For a given observation or target time (gray circles), the corresponding GFS value is selected from the most recent available forecast cycle based on its valid time. 
 The colored markers indicate which GFS forecast (by initialization time and lead time) is used to match each observation. 
 To construct a continuous sequence of predictions up to a 48-hour horizon, forecasts from eight consecutive GFS cycles are combined, ensuring that each observation uses the most temporally consistent forecast available.}
 
 \label{fig:gfs_schema}
\end{figure}

\begin{figure}[htbp]
 \centering
 
 \begin{subfigure}[b]{\linewidth}
 \centering
 \includegraphics[width=\linewidth]{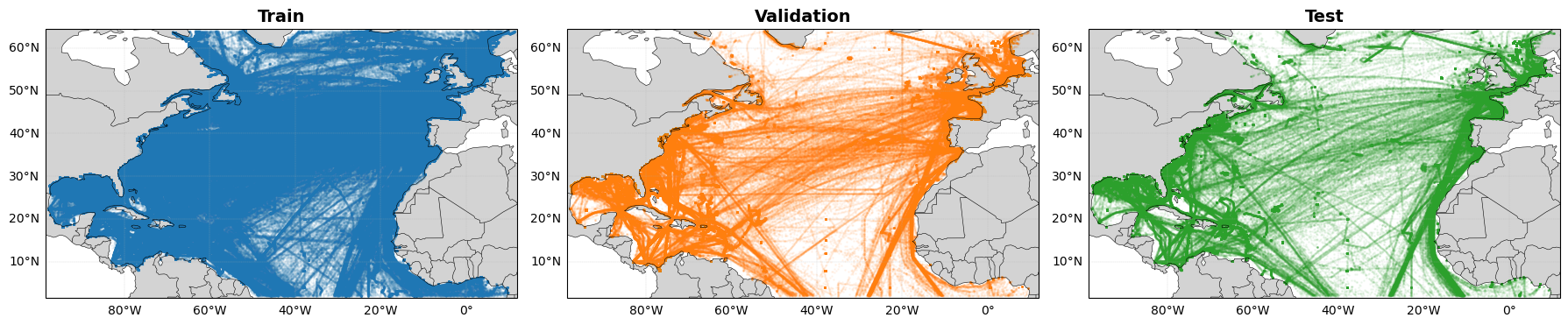}
 \caption{Spatial distribution of the training (blue), validation (orange), and test (green) subsets across the North Atlantic. Each color represents the observation locations assigned to the corresponding subset, ensuring spatial diversity and minimal overlap between regions.}
 
 \label{fig:spatial_split:top}
 \end{subfigure}
 
 \vspace{1em} 
 
 \begin{subfigure}[b]{\linewidth}
 \centering
 \includegraphics[width=\linewidth]{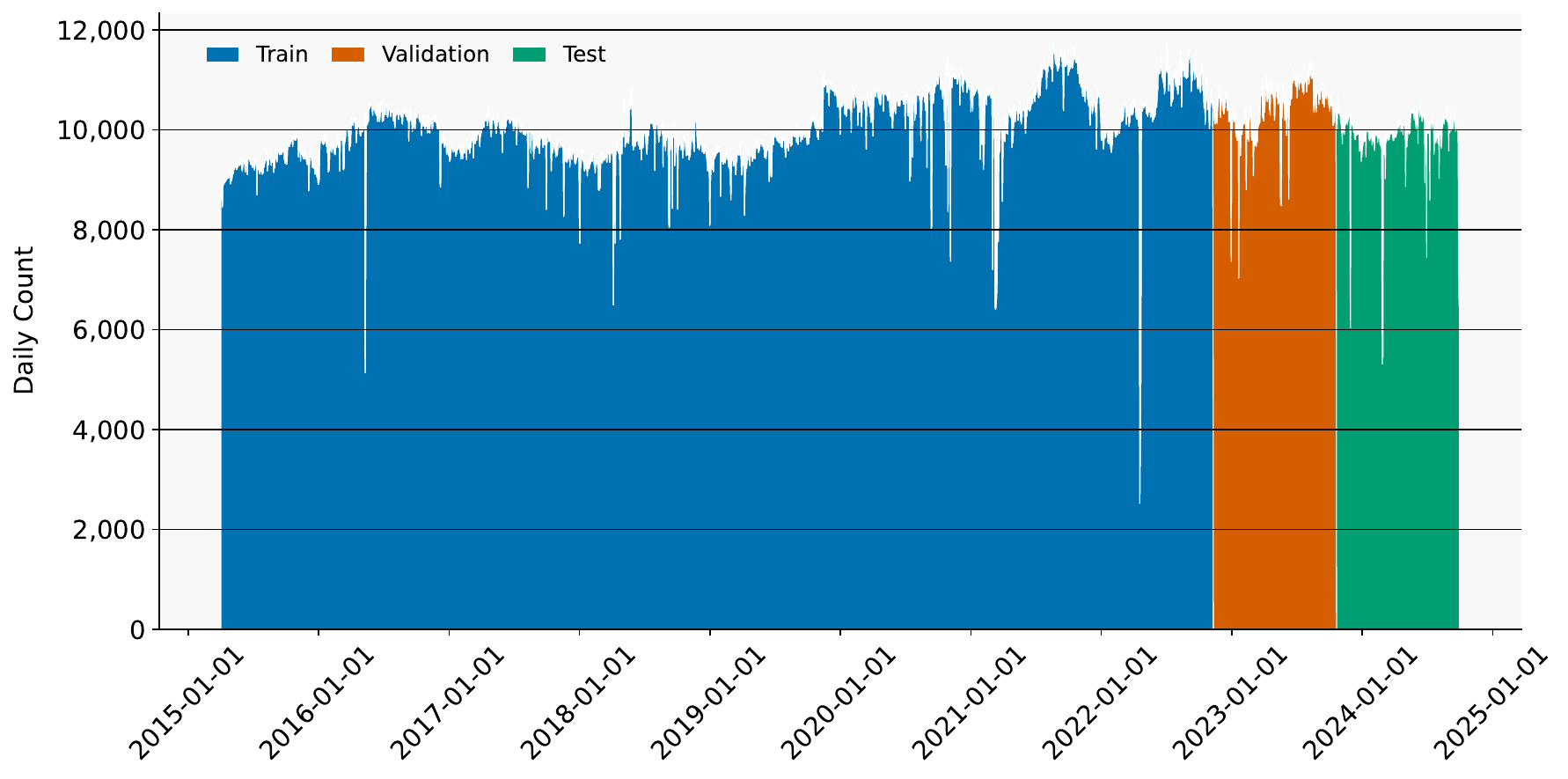}
 \caption{Temporal allocation of samples showing the chronological split between training, validation, and test periods. The temporal segmentation ensures that validation and testing occur strictly after the training period, avoiding temporal leakage.}
 \label{fig:time_split:bottom}
 \end{subfigure}
 
 \caption{
 Spatial and temporal partitioning of the dataset used for model development and evaluation. 
 The top panel illustrates the spatial separation between subsets, while the bottom panel shows the chronological division of samples through time. 
 Together, these splits ensure that the model is trained and evaluated on independent temporal domains.}
 \label{fig:set_split}
\end{figure}

\begin{figure}[h]
 \centering
 \includegraphics[width=1\linewidth]{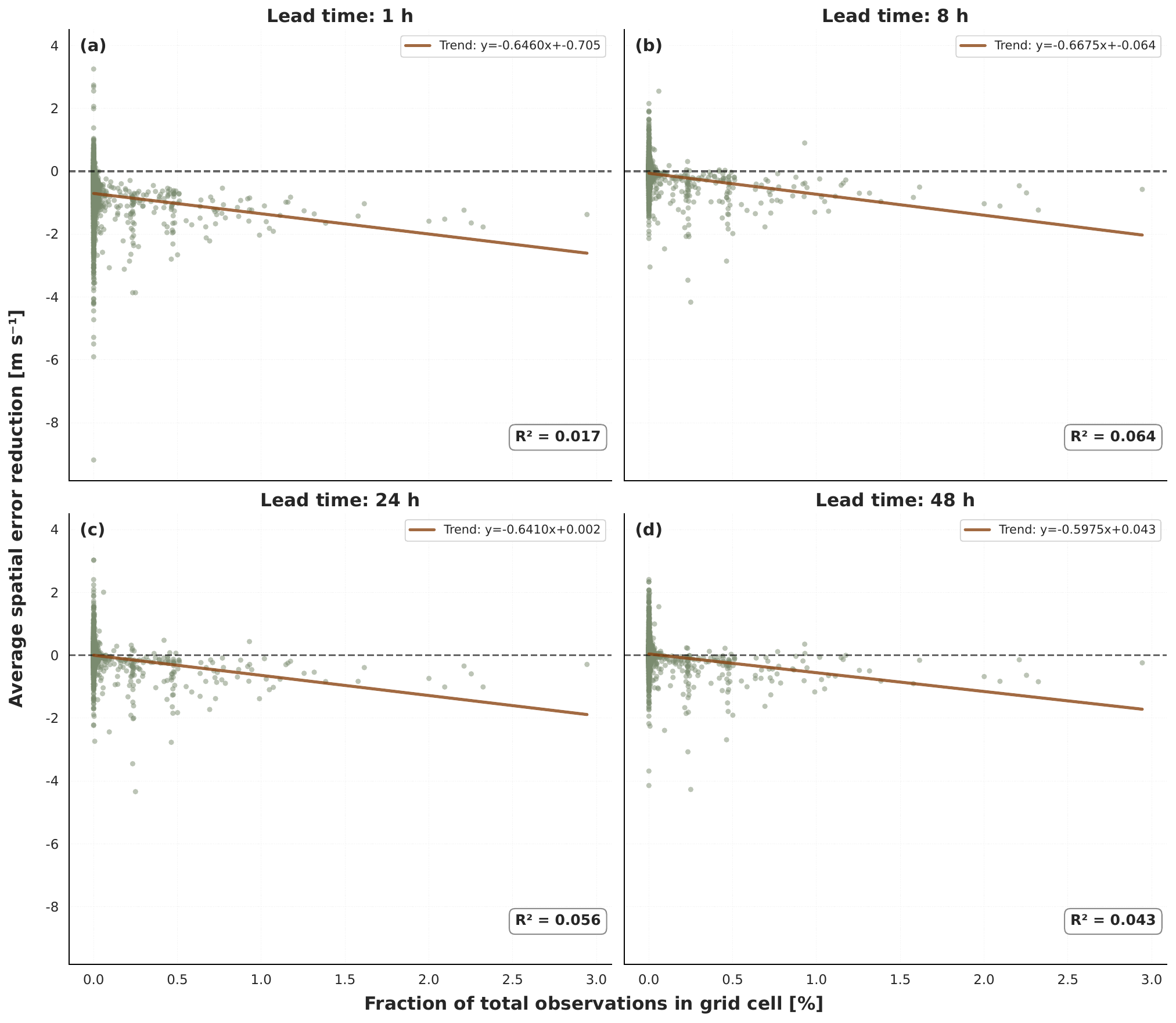}
 \caption{
 Relationship between observation density and spatial error reduction across lead times. Scatter plots showing the average spatial error reduction (model minus GFS) as a function of the fraction of total observations within each 1$^\circ$ $	imes$ 1$^\circ$ grid cell for lead times of (a) 1 h, (b) 8 h, (c) 24 h, and (d) 48 h. Each point represents a single grid cell, with negative values indicating that ORCA outperforms GFS. The dashed horizontal line at y = 0 marks the threshold where ORCA and GFS have equal performance. Linear trend lines (brown) indicate a consistent negative relationship between observation density and error reduction across all lead times. The trend suggests that grid cells with higher observation density tend to show larger improvements over GFS. Conversely, data-sparse regions show more variable performance, with some cells achieving substantial improvements (up to 4 m s$^{-1}$ error reduction) while others show degradation. 
}
 
 \label{fig:density_improvement}
\end{figure}

\begin{figure}[h]
 \centering
 \includegraphics[width=\linewidth]{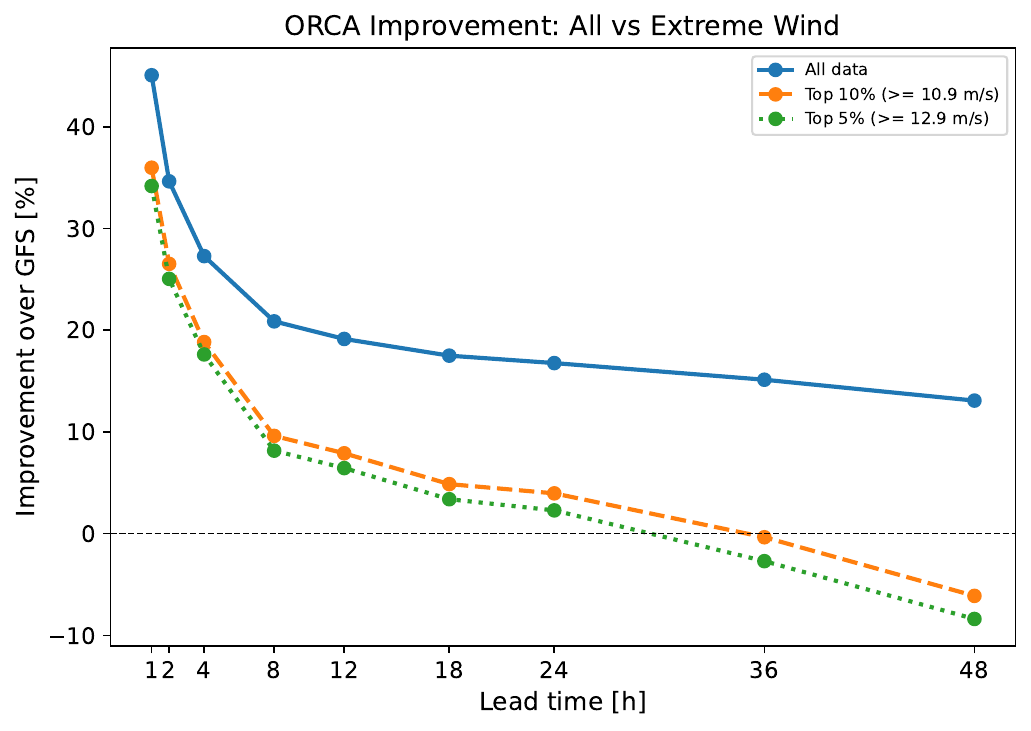}
 \caption{Relative improvement of ORCA over GFS as a function of lead time, stratified by observed wind-speed regime: all observations (blue, solid), top 10\% ($\geq$10.9~m\,s$^{-1}$, orange, dashed), and top 5\% ($\geq$12.9~m\,s$^{-1}$, green, dotted). The dashed horizontal line marks zero improvement. ORCA improves over GFS across all regimes at short lead times, but gains diminish faster for high-wind events and become negative beyond approximately 24~h for the top 10\% and 18~h for the top 5\%.}
 \label{fig:extreme_wind}
\end{figure}

\begin{figure}[h]
 \centering
 \includegraphics[width=\linewidth]{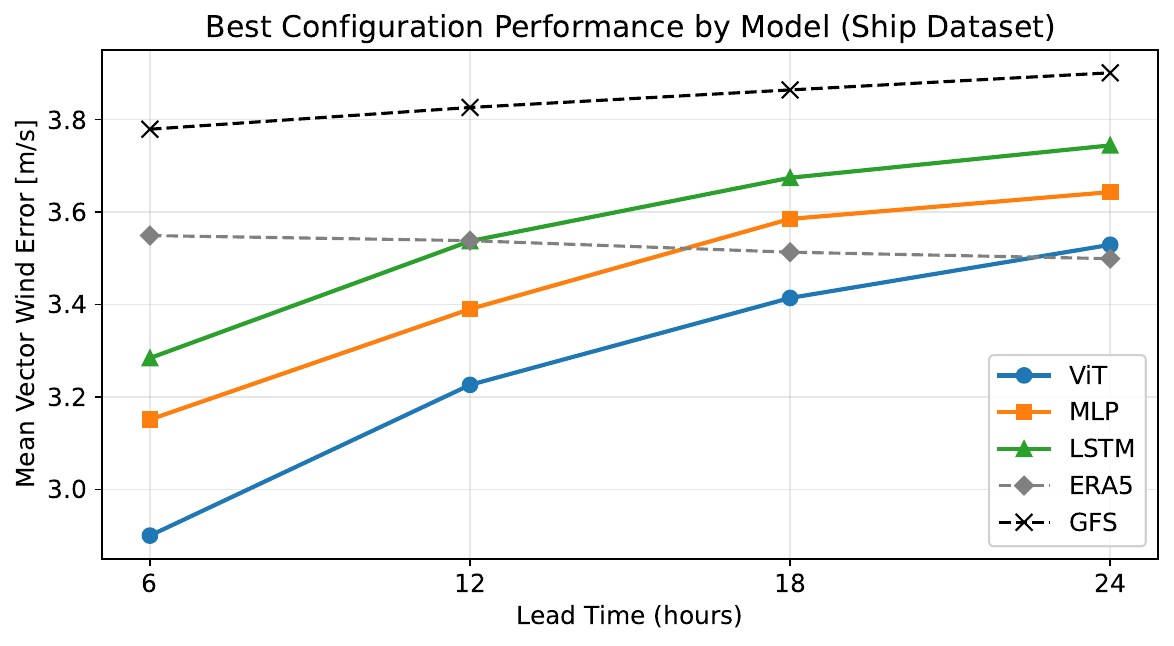}
 \caption{Comparison of best-configuration performance for MLP, LSTM, and ViT architectures on a ship-only subset of the ICOADS dataset (6-hour observation intervals, lead times 6--24~h). The ViT consistently outperforms both MLP and LSTM across all lead times, while all three models improve over the GFS baseline. ERA5 reanalysis is shown as a reference benchmark.}
 \label{fig:baseline_comparison}
\end{figure}

\subsection{Linear Regression Baseline}
\label{sec:lin_reg_baseline}

For each lead time, two independent ordinary-least-squares regressions
predict the target $u$ and $v$ components from a feature vector 
$\mathbf{x} \in \mathbb{R}^{12}$ comprising: (i)~the GFS forecast at the 
target location $(u^{\text{GFS}}_t, v^{\text{GFS}}_t)$, (ii)~target 
coordinates $(\phi_t, \lambda_t)$, (iii)~cyclical time features, 
(iv)~masked spatial means of nearby observations and co-located GFS 
values, and (v)~inverse-distance-weighted (IDW) analogues of~(iv), which vary per target and encode 
local spatial structure. The regression is trained on the same 
chronological split and evaluated under the identical masking and 
error metric as ORCA.

\begin{table}[htbp]
 \centering
 \caption{Mean vector wind error (m\,s$^{-1}$) for MLP, LSTM, and ViT (best configuration each) on the ship-only ICOADS subset (SD6). GFS and ERA5 are shown as baselines. Bold indicates the best-performing model per lead time.}
 \begin{tabular}{c S[table-format=1.3] S[table-format=1.3] S[table-format=1.3] S[table-format=1.3] S[table-format=1.3]}
 \toprule
 \textbf{Lead time} & {\textbf{MLP}} & {\textbf{LSTM}} & {\textbf{ViT}} & {\textbf{ERA5}} & {\textbf{GFS}} \\
 (h) & {(m\,s$^{-1}$)} & {(m\,s$^{-1}$)} & {(m\,s$^{-1}$)} & {(m\,s$^{-1}$)} & {(m\,s$^{-1}$)} \\
 \midrule
 6 & 3.151 & 3.284 & \textbf{2.900} & 3.549 & 3.779 \\
 12 & 3.390 & 3.537 & \textbf{3.226} & 3.538 & 3.826 \\
 18 & 3.585 & 3.674 & \textbf{3.414} & 3.513 & 3.864 \\
 24 & 3.643 & 3.744 & \textbf{3.529} & 3.499 & 3.901 \\
 \bottomrule
 \end{tabular}
 \label{tab:baseline_comparison}
\end{table}

\end{document}